# Scene restoration from scaffold occlusion using deep learning-based methods

Yuexiong Ding, Muyang Liu, Xiaowei Luo[*]

Department of Architecture and Civil Engineering, City University of Hong Kong, Hong Kong, China

xiaowluo@cityu.edu.hk

**Abstract.** The occlusion issues of computer vision (CV) applications in construction have attracted significant attention, especially those caused by the wide-coverage, crisscrossed, and immovable scaffold. Intuitively, removing the scaffold and restoring the occluded visual information can provide CV agents with clearer site views and thus help them better understand the construction scenes. Therefore, this study proposes a novel two-step method combining pixel-level segmentation and image inpainting for restoring construction scenes from scaffold occlusion. A low-cost data synthesis method based only on unlabeled data is developed to address the shortage dilemma of labeled data. Experiments on the synthesized test data show that the proposed method achieves performances of 92% mean intersection over union (MIoU) for scaffold segmentation and over 82% structural similarity (SSIM) for scene restoration from scaffold occlusion.

## 1. Introduction

A growing number of intelligent agents based on computer vision (CV) have been developed to replace the workforce for construction automation (Paneru and Jeelani, 2021). However, those existing CV-based models usually require a clear view without occlusion, or their performance might be seriously affected, resulting in biased results/decisions. Therefore, visual occlusion in construction has become a thorny issue and caught great research attention. Scaffold occlusion is one of the prominent issues since it is generally crisscrossing, wide-span, and can not be moved physically to alleviate occlusion. These natural characteristics make scaffold occlusion ubiquitous once construction scenes contain any scaffold, limiting CV-based models' performance in such circumstances. However, scaffold occlusion has not yet been studied, which may be due to the following challenges. First, start-up research from a blank field is generally tricky since there are few related works and datasets for reference. Besides, a pixel-level mask indicating the scaffold is required to process only the specified parts. However, it might take an unimaginable workforce and time to annotate each tubular of the whole scaffold at pixel level from an image because of its crisscrossing complex net structure. Finally, the scaffold usually occludes different targets or backgrounds due to its attribute of wide span. Therefore, inpainting and connecting objects and backgrounds with diverse textures and shapes in a natural transition will be another challenge.

This study proposes a two-step method that combines two deep learning-based models of scaffold segmentation and mask-aware image inpainting for scaffold occlusion removal. Specifically, a low-cost data synthesis method is first developed to generate sufficient labeled data using only unlabeled images to facilitate model training and evaluation. Then a mask image indicating the scaffold pixels is automatically identified by a pixel-level segmentation model from the scaffolding image. Finally, a mask-aware inpainting model is applied to restore the occluded regions precisely based on the image context, which can handle the transition restoration between different textures well. The novelties and contributions of the study can be summarized as follows: 1) a new two-step method is proposed for precise localization and restoration of scaffold occlusion; 2) A low-cost data synthesis method is proposed for labeled data generation, which significantly alleviates the shortage dilemma of related labeled data in construction; 3) The proposed two-step method achieves better performance than the existing img2img translation method. The rest of the paper is organized as follows: Section 2 reviews relevant research and methods. Section 3 describes the overall research framework, and Section 4 conducts experiments to verify the effectiveness of the proposed two-step method. Section 5 concludes the study and its limitations.

## 2. Literature review

### 2.1 Occlusion-related research in construction

Most of the occlusion-related works in construction took occlusion as an experimental variable to discuss the robustness of the model by comparing the performance with and without occlusion or with different proportions of occlusion. For example, by setting different degrees of occlusion, Chen et al. (2018) found that reducing occlusion can significantly improve the accuracy of 3D object recognition from the point cloud data collected on-site. However, no specific solutions to occlusion issues were proposed after discussing and verifying the impact of the occlusion. Some research attempted to propose specific methods to solve occlusion issues. To solve occlusion, Li et al. (2022) developed a skeleton-based worker action recognition model under occlusion conditions, which applied a generative adversarial network (GAN) to perform numerical interpolation to predict the occluded skeleton of workers without removing the obstacles. However, restoring the occluded visual information of a construction scene is more basic than restoring the occluded skeleton of workers since the restored scene can serve more downstream CV tasks. Therefore, Angah and Chen (2020) removed the workers in front of the building by developing an img2img translation model based on a convolutional neural network (CNN) and GAN to restore the occluded visual information. Though the img2img translation strategy provides a one-step solution for occlusion removal, it modifies the whole image rather than processing only the occluded regions and thus leads to more bias errors. Restoring only the occluded areas might be more reasonable since it preserves as much original visual information as possible.

### 2.2 Deep learning technologies for scaffold occlusion removal

The proposed two-step method mainly involves two deep learning-based CV tasks: semantic segmentation and image inpainting. Separating the pixel-level scaffold from an image in the first step is a typical semantic segmentation task that classifies pixels into the instance categories to which they belong. In other words, semantic segmentation is a pixel-level classification task. Almost all semantic segmentation models used the CNN as the backbone over the past few years, such as the U-Net (Ronneberger et al., 2015), DeepLab (Chen et al., 2017), and FastFCN (Wu et al., 2019) models. The Transformer is an emerging deep backbone network based on the self-attention mechanism, which was first proposed and used in natural language processing (NLP) tasks for language representation learning (Vaswani et al., 2017). When extended to the CV field, the Transformer has achieved state-of-the-art (SOTA) performance in many downstream CV tasks (Dosovitskiy et al., 2021), including the semantic segmentation task (Liu et al., 2021). Image in sequence format and the self-attention mechanism provide Transformer-based models with a larger receptive field than CNN architectures, which is more conducive to being applied to segment the crisscrossing and wide-span scaffold.

Mask-aware image inpainting is another technology adopted in this study, which tries to restore the missing regions of the image based only on the visual information of the known areas. The GAN opens a door for image inpainting since it can generate images that confuse the real with the fake and deceive human eyes. The GAN comprises two independent deep neural networks competing with each other, named generator and discriminator (Goodfellow et al., 2014). The generator tries to generate artificial images as naturally as possible, while the discriminator attempts to discriminate artificial images from real ones. Through the iterative adversarial competition, the two networks' performances are alternately getting better. The mainstream models nowadays for image inpainting currently still adopt the GAN architecture with multiple CNN layers (Yi et al., 2020; Yu et al., 2019; Zeng et al., 2021). The CNN here refers to not only the traditional one but also the Gated CNN, Dilated CNN, and their combination (Zeng et al., 2021). Due to the crisscrossing and wide-spanning, repairing the scaffold occlusion needs to focus on natural transition when restoring missing areas across multiple objects and backgrounds. Some context-based models (Yi et al., 2020; Zeng et al., 2021) that focus on the transition effect between known and missing regions thus are the primary references to adopt and train the inpainting model in this study.

## 3. Method

As mentioned above, restoring only the occluded regions to preserve as much original visual information as possible might be more reasonable than what the existing img2img method has done. A pixel-level mask of the scaffold thus should be provided for the restoration model to indicate the occluded parts. Guided by this idea, a two-step method is designed in this study, which first recognizes the scaffold at the pixel level to generate an indication mask and then restores the regions indicated by the mask based on the contextual unoccluded information. Figure 1 shows the research framework for constructing the designed two-step method, mainly including three parts: 1) data synthesis, 2) pixel-level scaffold segmentation modeling, and 3) context-based image inpainting modeling. Finally, the two-step method for the removal of scaffold occlusion is realized by sequentially executing the well-trained segmentation and inpainting models

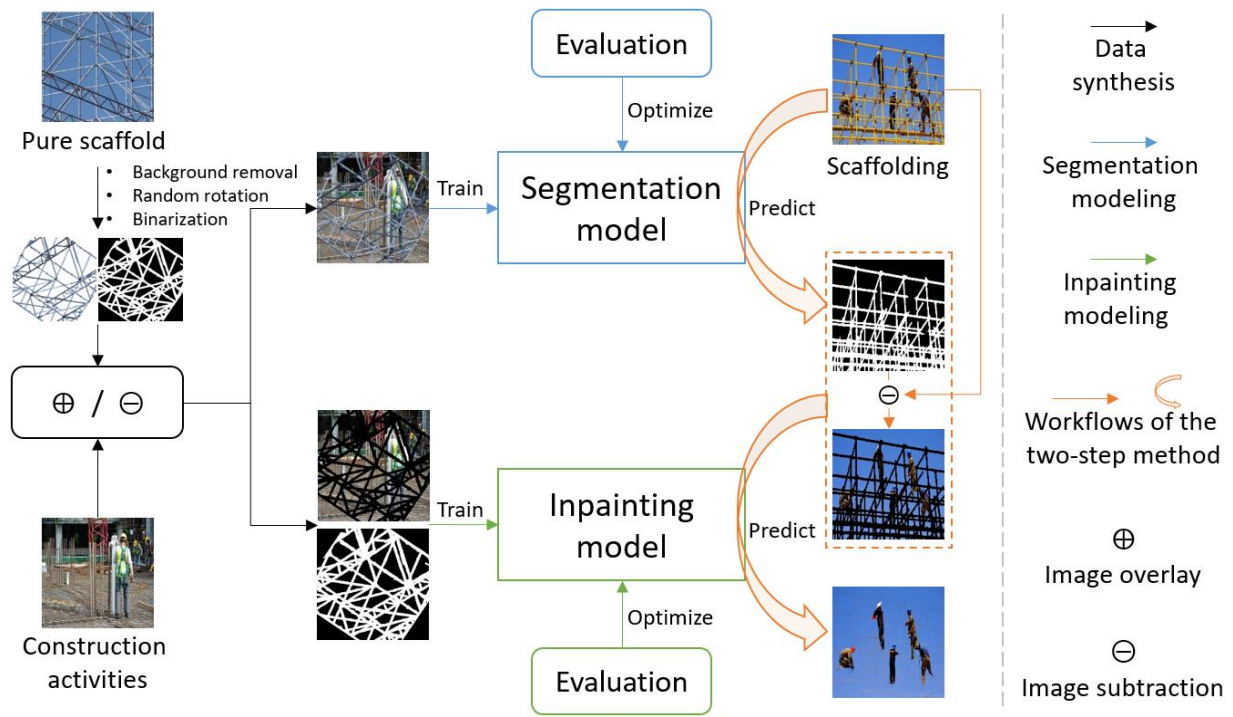

Figure 1. The research framework for constructing the proposed two-step method.

## 3.1 Data synthesis

As shown in Figure 1, the labeled data was automatically synthesized in the following three steps. First, the backgrounds of the collected pure scaffold images were removed using background removal technology (Removal AI, 2019), which realized foreground and background separation by identifying foreground targets (not paying attention to target categories). Note that a pure background must be strictly met in this stage, or the Removal AI can not recognize and separate the scaffold. Those separated scaffolds were then rotated randomly to achieve better diversity and binarized to generate pixel-level mask maps. Finally, image overlay and subtraction operations were applied to the collected construction activity images using the rotated and binary results. The final overlaid and subtracted images in this study were both $M \times N$, where $M$ and $N$ are the numbers of pure scaffold and construction activity images. Sufficient labeled data thus become easily obtained from a small amount of collected unlabeled data.

## 3.2 Pixel-level scaffold segmentation

A typical semantic segmentation model generally comprises a backbone network and a segmentation head. Considering the encouraging performance of Swin-Transformer (Liu et al., 2021) and UperHead (Xiao et al., 2018) in many CV tasks, these two networks were adopted in this study as the backbone and segmentation head to build the pixel-level scaffold segmentation model. Figure 2 shows some core components of the scaffold segmentation model. A series of Swin-Transformer blocks connect in a hierarchical approach to obtain different resolution feature maps after each stage, as $S_1, S_2, S_3, S_4$ shown in Figure 2 (a). The backbone size can be easily controlled by setting the blocks number ($N_i$) and the feature dimension ($C$) in each stage. The Swin-Transformer is characterized by alternating use the windows multi-head self-attention (W-MSA) and shifted W-MSA (SW-MSA), ensuring the network can still obtain the global receptive field while reducing the calculation workload, as shown in Figure 2 (b) and (c). Four new feature maps ($P_1, P_2, P_3$, and $P_4$) are then generated from the feature maps extracted by the backbone ($S_1, S_2, S_3$, and $S_4$) using the feature pyramid network (FPN) and pyramid pooling module (PPM) (Xiao et al., 2018) and are further fused into a unified feature map to complete the final scaffold segmentation, as shown in Figure 2 (d).

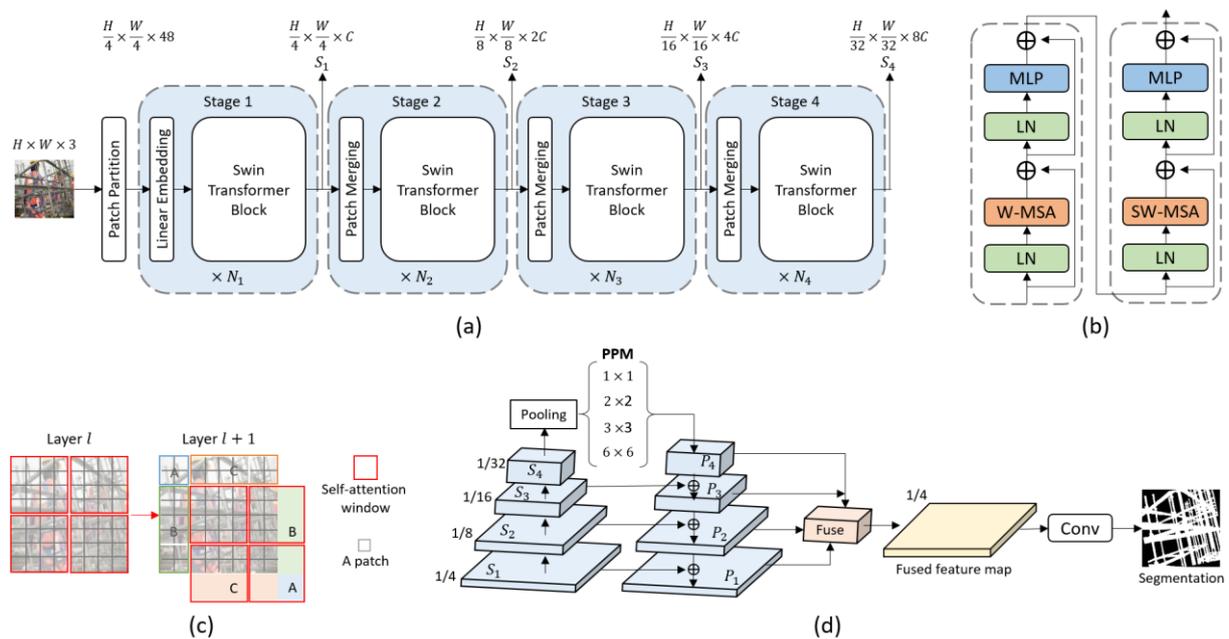

Figure 2. (a) The backbone network of the scaffold segmentation model; (b) The internal structure of two successive Swin-Transformer blocks; (c) The shifted window approach for computing self-attention; (d) The architecture of the scaffold segmentation head.

## 3.3 Context-based image inpainting

A mask-aware inpainting model with contextual reconstruction (CR) capability was built in this study by referring to the CR-Fill (Zeng et al., 2021). The CR-Fill is also a GAN model characterized by its generator comprising the coarse and refinement networks with the special CR loss ($\mathcal{L}_{CR}$), as shown in Figure 3. The main idea of the CR loss is to reconstruct patches containing missing pixels in a feature map by weighted averaging feature patches from the completely known regions. Specifically, the feature map from the generator is first input into a similarity encoder to obtain a similarity matrix ($S$) of different types of patches, which can be represented as $S_{ij} = \frac{s(u_i)^T s(u_j)}{\|s(u_i)\| \cdot \|s(u_j)\|}$, where $u_i$ denotes patches containing missing pixels, $u_j$ denotes patches from known regions, and $S_{ij}$ is the similarity between patch $u_i$ and $u_j$. Then patches ($f(u_i)$) containing missing pixels in the new feature map ($F(u)$) generated by the auxiliary encoder from the raw occluded image are replaced via weighted average: $\bar{f}(u_i) = \sum_{j \in V} \text{softmax}(\alpha S_{ij}) f(u_j)$, where $V$ is the index set of patches from known regions. After patch replacement, the reconstructed feature map ($F(u)$) is decoded into an auxiliary image by the auxiliary decoder, which learns lazily to copy the input to the output to make the auxiliary decoder invert the auxiliary encoder easily, i.e., $A(U) = H(F(U)) = U$, where $U$ is the ground-truth image, $A(\cdot)$ represents the auxiliary image, and $H(\cdot)$ is the auxiliary decoder. Finally, the inpainting loss of the auxiliary image is a sum of the local loss of patches: $\mathcal{L}(A(U)) = \sum_{i \in V'} l_i(a(\bar{f}(u_i)))$, where $V'$ is the index set of patches containing missing pixels.

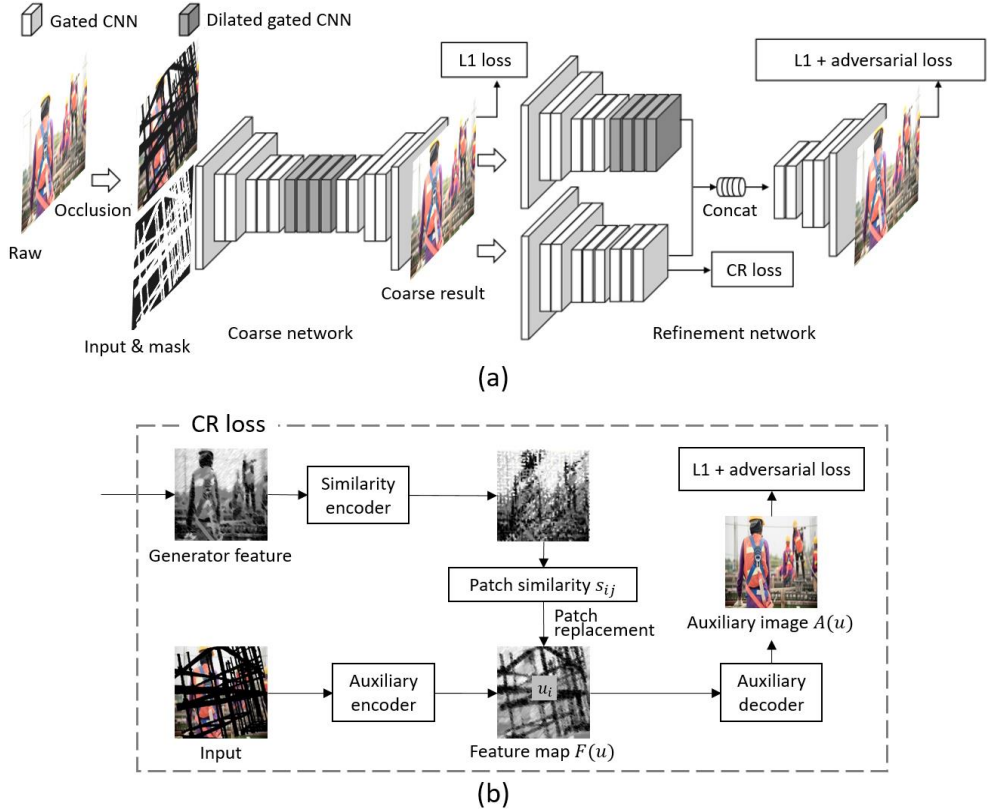

Figure 3. (a) The generator network of the inpainting model; (b) Details of the CR loss module.

## 4. Experiments

### 4.1 Data preparation

Three types of images with a total number of 930 were collected from the Internet for experiments, including 230 pure scaffold images and 700 construction activities images, as shown in Figure 4. Then a total number of 120,000 images for each task were synthesized according to the method described in section 3.1 using 200 pure scaffold images and 600 construction activities images, which were further shuffled and divided into training (*syn_train*), validation (*syn_val*), and test (*syn_test*) sets. Besides, an extra dataset (*syn_ext_test*) of 3,000 images was synthesized using the remaining 30 pure scaffold images and 100 construction activities images to validate the generalization of the two trained models for totally new images. Table 1 shows the statistics information by different scaffold proportions (also called missing rate in the inpainting task) for each sub-dataset.

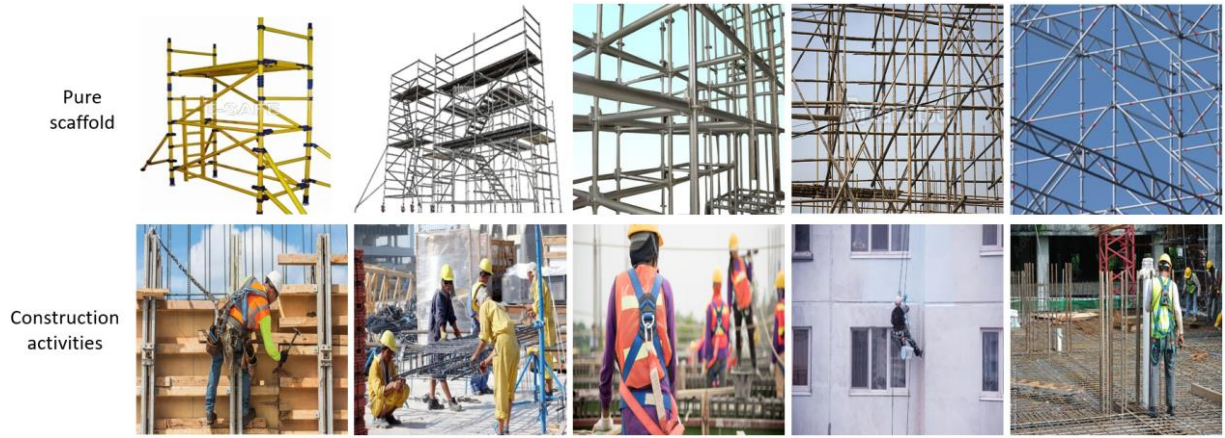

Figure 4. Samples of the collected images.

Table 1. Synthesized image numbers in different scaffold proportions.

| Dataset | (0, 0.2] | (0.2, 0.4] | (0.4, 0.6] | (0.6, 0.8] | (0.8, 1.0) | **Total** |
|---|---|---|---|---|---|---|
| *syn_train* | 2080 | 47,152 | 43,976 | 14,724 | 68 | 108,000 |
| *syn_val* | 109 | 2631 | 2455 | 796 | 9 | 6,000 |
| *syn_test* | 131 | 2584 | 2449 | 830 | 6 | 6,000 |
| *syn_ext_test* | 100 | 2257 | 643 | 0 | 0 | 3,000 |

* (0, 0.2] denotes the number of images with scaffolding proportions between 0 and 0.2.

### 4.2 Scaffold segmentation

The backbone of the scaffold segmentation model in this study was customized with the configurations of [2, 2, 18, 2] for block number ($N_i$) in each stage, 128 for feature dimension ($C$), and 7 for the local self-attention window size. Then the pre-trained weights of the Swin-B (Liu et al., 2021) were utilized to initialize the backbone. Finally, the scaffold segmentation model was trained on the synthesized data using the following settings: the Adamw optimizer with a learning rate of 1e-6, the image size of 512 × 512, the batch size of 2, and the total training iteration of 160K. The mean intersection over union (MIoU) was adopted for scaffold segmentation evaluation, which avoids biased evaluation caused by different category shape sizes by calculating the IoU of the predictions ($P_i$) and ground truths ($GT_i$): MIoU = $\frac{1}{K}\sum_{i=1}^{k}\frac{P_i \cap GT_i}{P_i \cup GT_i}$, where $K$ is the number of pixel categories. Higher MIoU means better segmentation performance.

This study identified two pixel categories: scaffold and background. Figure 5 shows the model performance for the two categories on the *syn_test* and *syn_ext_test* datasets. The trained segmentation model achieves an encouraging performance of 92% MIoU on datasets *syn_test* and *syn_ext_test*. Besides, the model performance on the *syn_ext_test* dataset is roughly the same as that on the *syn_test* dataset, indicating the excellent generalization of the trained model for new images. Good model generalization also proves the feasibility and effectiveness of the developed data synthesis method and its synthesized data. Finally, the model performance (MIoU) gradually declines with increased scaffold proportion, which reveals that it is more difficult for the model to segment a larger scaffold from an image. A balanced performance is achieved when the proportions of two categories in an image are roughly similar.

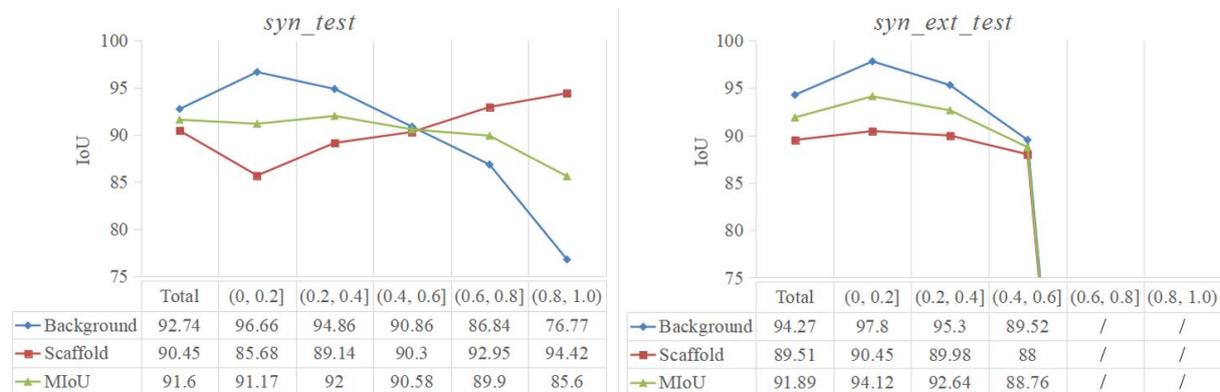

Figure 5. Scaffold segmentation performances on datasets *syn_test* and *syn_ext_test*.

## 4.3 Image inpainting for scaffold occlusion removal

The inpainting model in this study was developed and initialized based on the CR-Fill (Zeng et al., 2021) and then was fine-tuned on the synthesized data with the following settings: the Adam optimizer with a learning rate of 1e-4, the image size of $512 \times 512$, the batch size of 4, and the total training iteration of 1M. For image inpainting evaluation, standard metrics include the mean absolute error (MAE), structural similarity (SSIM), peak signal-to-noise ratio (PSNR), Fréchet inception distance (FID). Lower MAE and FID but higher SSIM and PSNR indicate better inpainting performance. Since the inpainting result is also the final result of the proposed two-step method, the inpainting performance can be regarded as the performance of scaffold occlusion removal. For comparison, the img2img translation model (Angah and Chen, 2020) was also trained (named Pix2pix_U-Net) and evaluated on two synthesized datasets, as shown in Table 2.

Compared with the img2img translation model, the proposed two-step method performs best on two test datasets, mainly due to its precise localization and restoration strategy of scaffold occlusions. Besides, large FID fluctuations of the two models might reveal a defect of the img2img translation strategy that the generation of the new image destroys the high-level image features of the unoccluded regions. Notably, the FID fluctuation is more significant than fluctuations of the MAE, SSIM, and PSNR of the same model between different test sets, indicating that the trained inpainting model has limited generalization in restoring high-level image features for new images. The problem may be caused by the training data rather than the models since the two trained models have the same phenomenon. In fact, only 600 original construction activities images were used for data synthesis, which can not perfectly construct a high-level image feature domain of the construction field, resulting in limited generalization in high-level feature restoration for new images. Finally, though the scaffold removal performance gradually declines with the missing rate (i.e., the scaffold proportion) increase, it keeps an encouraging performance of at least 0.8 SSIM when the missing rate is less than 0.6. However, there is a sharp performance decline when the missing rate is over 0.8, where the restored images are only 53% similar (SSIM) to ground truth images.

Table 2. Scaffold removal performances on the *syn_test* and *syn_ext_test* datasets.

| Model | Dataset | Missing rate | MAE | SSIM | PSNR | FID |
| --- | --- | --- | --- | --- | --- | --- |
| Pix2pix_U-Net | *syn_test* | Total | 0.0692 | 0.64 | 18.74 | 90.35 |
|  | *syn_ext_test* | Total | 0.0726 | 0.63 | 18.45 | 154.91 |
| The proposed method | *syn_test* | Total | 0.0324 | 0.82 | 22.77 | 35.78 |
|  | *syn_ext_test* | Total | 0.0253 | 0.86 | 24.10 | 62.80 |
|  | *syn_test* | (0, 0.2] | 0.0116 | 0.94 | 28.78 | 24.00 |
|  |  | (0.2, 0.4] | 0.022 | 0.88 | 24.74 | 29.22 |
|  |  | (0.4, 0.6] | 0.0358 | 0.80 | 21.79 | 45.99 |
|  |  | (0.6, 0.8] | 0.053 | 0.71 | 19.53 | 78.62 |
|  |  | (0.8, 1.0) | 0.0791 | 0.53 | 17.62 | 226.66 |

**4.4 Real-world application**

More field tests were conducted on some real-world scaffolding images to verify the feasibility and practicability of the proposed two-step method, as shown in Figure 6. Note that the performance evaluation in this stage was based only on human-eye observation instead of metrics since it is almost impossible to obtain ground-truth scaffold removal results for these scaffolding images. The img2img method (Pix2pix_U-Net) mentioned above was adopted for direct comparison to make observation-based evaluation more convincing. Compared with the img2img method, the proposed method achieves better results in scaffold occlusion removal, mainly reflected in accurately recognizing and removing almost every scaffold tube. On the contrary, it can be clearly observed with the naked eye that many scaffold tubes are left in the img2img results. Besides, the proposed method only restores the occluded regions in the second step benefitting from precise scaffold segmentation in the first step, thus retaining more image's original information than the img2img method.

**5. Conclusions**

This study proposes a deep learning-based two-step method for scaffold occlusion removal, combining data synthesis, semantic segmentation, and image inpainting technologies. Based on the developed data synthesis method, sufficient labeled data are synthesized from the collected unlabelled images to support the supervised learning of the segmentation and inpainting models. Experiments on the synthesized data show that the pixel-level scaffold segmentation model achieves an excellent performance of roughly 92% MIoU, and the final scaffold occlusion removal results are more than 82% similar (SSIM) with the ground truths. This study might have great significance to the relevant fields since it is the first attempt to explore the feasibility of scene restoration from scaffold occlusion. Besides, the proposed method can provide clearer views for many downstream CV tasks in scaffolding scenarios to help achieve better performance. Future efforts can focus on enlarging the image representation domain and compressing the two-step method into a multi-task end-to-end lightweight model to meet the real-time requirement.

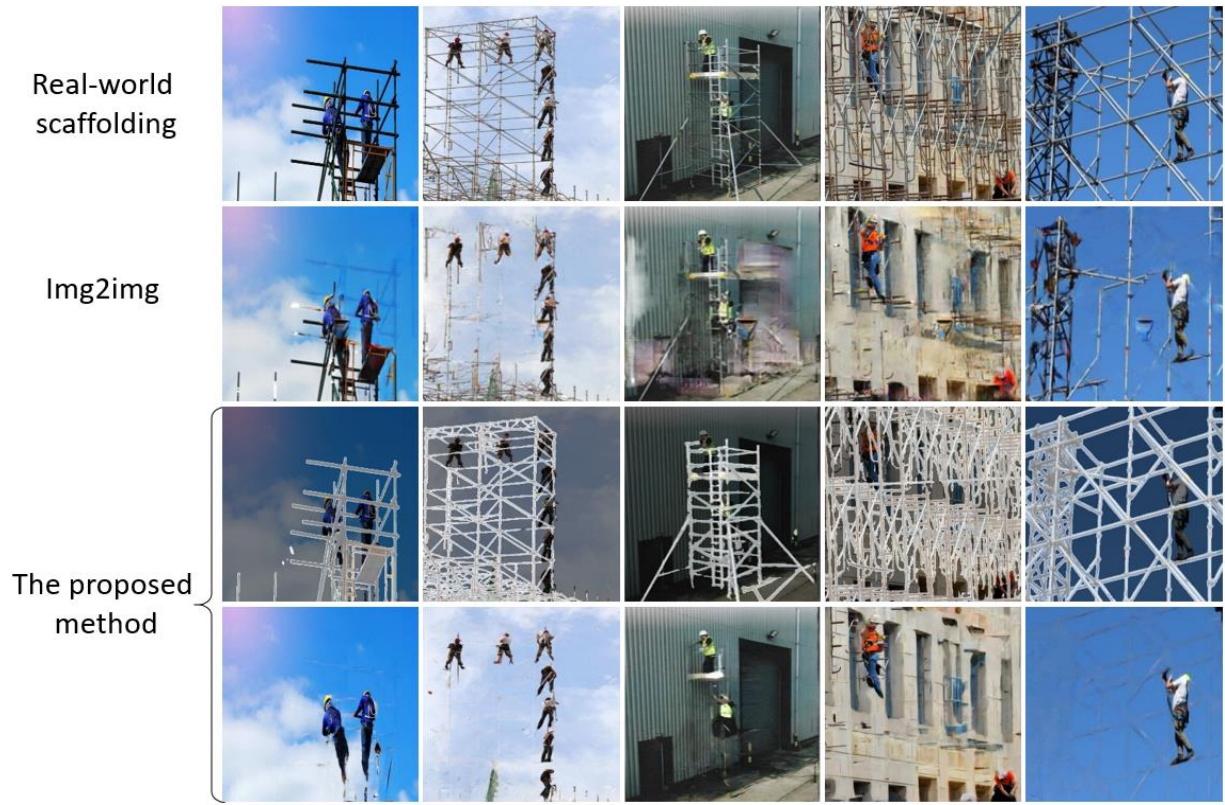

Figure 6. Real-world scaffold occlusion removal. The amplified details from left to right are cut from real-world scaffolding, results of the img2img method, and results of the proposed method.